# Rethinking Deep Learning:
# Propagating Information in Neural Networks
# without Backpropagation and Statistical Optimization

Kei Itoh*

August 18, 2024

## Abstract

Developing strong AI signifies the arrival of technological singularity, contributing greatly to advancing human civilization and resolving social issues. Neural networks (NNs) and deep learning, which utilize NNs, are expected to lead to strong AI due to their biological neural system-mimicking structures. However, the statistical weight optimization techniques commonly used, such as error backpropagation and loss functions, may hinder the mimicry of neural systems. This study discusses the information propagation capabilities and potential practical applications of NNs as neural system mimicking structures by solving the handwritten character recognition problem in the Modified National Institute of Standards and Technology (MNIST) database without using statistical weight optimization techniques like error backpropagation. In this study, the NNs architecture comprises fully connected layers using step functions as activation functions, with 0-15 hidden layers, and no weight updates. The accuracy is calculated by comparing the average output vectors of the training data for each label with the output vectors of the test data, based on vector similarity. The results showed that the maximum accuracy achieved is around 80%. This indicates that NNs can propagate information correctly without using statistical weight optimization. Additionally, the accuracy decreased with an increasing number of hidden layers. This is attributed to the decrease in the variance of the output vectors as the number of hidden layers increases, suggesting that the output data becomes smooth. This study's NNs and accuracy calculation methods are simple and have room for various improvements. Moreover, creating a feedforward NNs that repeatedly cycles through 'input → processing → output → environmental response → input → ...' could pave the way for practical software applications.

* Ehime University Graduate School of Science and Technology. The author conducted this research independently while affiliated with Ehime University, and utilized the university's library and electronic resources. The research was conducted without specific financial support from external funding sources. The author declares no conflict of interest.

## 1. Introduction

It is well-known that one of the major ways to reach the technological singularity is through the development of strong AI (general-purpose AI) [e.g., 1,2]. Strong AI, by becoming the driving force behind civilizational progress, could solve various problems and achieve numerous goals within our civilization [e.g., 3]. One specific method for realizing strong AI that has been attempted for a long time is mimicking biological neural systems [e.g., 4,5]. Many multicellular organisms, including humans, possess a nervous system that transmits information through electrical signals. Learning and memory occur by altering the strength and structure of connections between neurons, the basic units of the nervous system, based on received information [e.g., 6]. Mimicking this has been applied to artificial intelligence development using neural networks (NNs) structures, originating from the McCulloch-Pitts model [4] and Rosenblatt's perceptron model [5]. There is hope that these technologies can bring us closer to the ultimate form of intelligence, akin to human intelligence.

The development of artificial intelligence using NNs and their learning methods currently relies fundamentally on error backpropagation [e.g., 7], making their capabilities practical. Additionally, since the backpropagation mechanism is rarely observed in biological neural systems, the development of NNs-based artificial intelligence that does not use error backpropagation has also been pursued [e.g., 8, 9]. However, both approaches essentially involve a statistical optimization process for weights using loss functions or fitness functions. It is considered that biological organisms do not perform statistical weight optimization, not just backpropagation. This is evident from the early achievements in neuroscience by Santiago Ramón y Cajal and Donald Olding Hebb [10,11]. Moreover, the goal of NNs-based artificial intelligence often centers on treating NNs as functions and optimizing them. From the perspective of mimicking neural systems, however, it should be interpreted that biological organisms possess algorithmic functionality and structure [e.g., 12,13,14]. Statistical weight optimization could lead to the rigidity of such algorithmic functions in NNs. Biological organisms perform input, output, and information processing under the influence of their surrounding environment [e.g., 15,16,17,18,19]. If one considers these functions as being performed by a function, the environment itself must be incorporated into the structure as a variable. Integrating loss functions or fitness functions as environmental variables is likely difficult (if not impossible), and it is clear that such functions are not currently suitable for handling diverse information.

Fundamentally, the very propagation of information, one of the principal abilities of neural systems, has likely not been sufficiently discussed in NNs development. This should be one of the most critical considerations when viewing NNs as a mimicry of neural systems, but as mentioned earlier, the importance of this aspect is diminished (or skipped) by weight updates through loss functions and backpropagation. While NNs exist as information in computer programs, neural systems exist as structures in biological organisms. For vertebrates to form their bodies and function according to genetic information, appropriate neural circuitry is essential [e.g., 6,19], indicating the importance of the information propagation capability of neural systems. This also highlights the structural deficiencies of current NNs compared to neural systems. From this, it is evident that current NNs-based artificial intelligence development theories lack the construction and analysis of NNs focusing solely on feedforward and excluding statistical weight optimization.

The purpose of this study is to demonstrate the information propagation capability of feedforward NNs constructed without statistical weight optimization, and to analyze and interpret this capability. Specifically, a simple fully connected neural networks with 0-15 hidden layers will be used to input the Modified National Institute of Standards and Technology database (MNIST) [20], and its accuracy of handwriting recognition will be calculated using a method that compares the average output vectors of training data and the output data of test data, based on the concept of distributed representation. Additionally, analyses will be conducted by calculating accuracy over a broader correct range and computing the variance of output vectors. From these results and analyses, interpretations will be provided regarding the information propagation capability of NNs, along with discussions on the structuring of NNs and their application to mimicking biological neural systems.

**2. Calculation Method**

In this study, NNs are constructed using the Pytorch library (ver. 2.2.1) in the Python language (ver. 3.10.11). The dimension of the input layer, hidden layers, and output layer are all unified to 784, and the number of hidden layers is set to 0-15 layers for verification. All NNs are composed of fully connected layers, and the step function is used as the activation function. The weights are sampled according to a uniform distribution from the range U(-sqrt(1/784), sqrt(1/784)), and no weight updates are performed. These simple settings are intended to focus on observing the information propagation capabilities of NNs.

The MNIST training data is input into the constructed NNs, and the average vector of the output vectors is calculated for each label from 0 to 9. The label with the highest vector similarity, calculated from the Euclidean distance between the average output vector of handwritten digits 0-9 and the output vector obtained by inputting the test data into the NNs, is taken as the predicted value. The accuracy rate is obtained by verifying whether the predicted value matches the correct labeling of the test data for all test data. This verification method is inspired by distributed representation [21, 22]. Typically, vector similarity in distributed representation is used to define the meaning of input vector data. In this study, distributed representation is applied to calculate the accuracy rate from the similarity between output vectors.

The simplified workflow of the accuracy calculation program in this study is as follows:

1. Importing Libraries: Import the necessary Python libraries (torch, torchvision, numpy, etc.).
2. Loading and Preprocessing MNIST Data: Scale the pixel values of the MNIST data to 0-1. Load the MNIST training and test data.
3. Defining Network Structure: Set up the fully connected layers and feedforward structure, using the step function as the activation function.
4. Initializing and Deploying the Model: Initialize the defined model and deploy it to the GPU.
5. Calculating Average Vectors: Define a function to calculate the average vector for each class using the training data. Input the training data into the network and execute this function on the output data to compute the average vector for each class.
6. Evaluating Test Data: Define a function for evaluating the test data by calculating the Euclidean distance between the output vector and the average vector. Input the test data into the network, compute the predicted value from the evaluation function on the output data, and calculate the accuracy rate.
7. Saving Results: Save various necessary information in a text file.

The computations in this study are performed using NVIDIA GeForce RTX 4060 Ti 16GB, which provides ample computational resources. The only computational cost that increases with the number of hidden layers is the forward propagation process, and the computation time remains almost the same for any number of hidden layers.

## 3. Results

Figure 1 shows the accuracy rates for different numbers of hidden layers in the NNs. Figure 1.A presents the accuracy rate when the predicted values perfectly matched the correct labels, while Figure 1.B displays the accuracy rate when the range of correct answers (prediction range) is expanded to the top three closest in vector similarity. For up to about 10 hidden layers, the accuracy rates are higher than those obtained by random guessing (Accuracy = 10% in Figure 1.A and 30% in Figure 1.B). Among these, with 0-2 hidden layers, the accuracy rate is approximately 80% in Figure 1.A and about 95% in Figure 1.B. However, increasing the number of hidden layers to 15 resulted in accuracy rates similar to random guessing. This trend persisted even when further increasing the number of hidden layers. These results suggest that even with a simple NNs structure using fully connected layers and step functions, it is possible to correctly propagate input information without backpropagation mechanisms or statistical weight optimization.

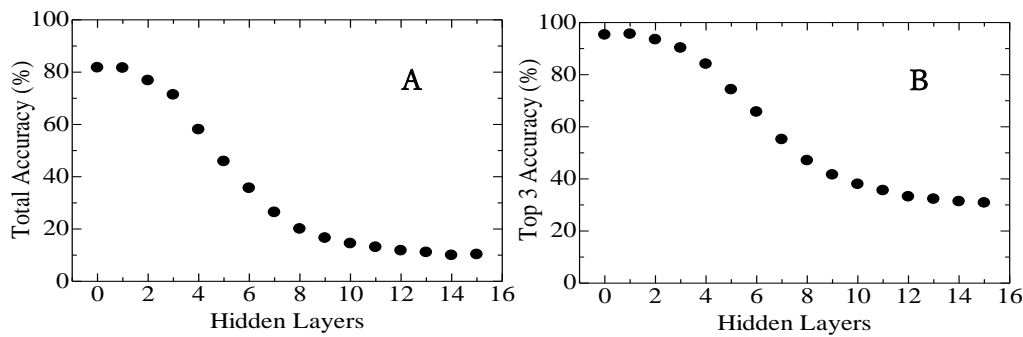

Figure 1: Accuracy rates for different numbers of hidden layers.

In traditional deep learning classification problems (such as MNIST), the output layer dimensions are typically matched to the number of correct labels, and the softmax function is used to compute the predicted values [e.g., 23]. Matching the dimensions to the number of labels and using the softmax function are also inconsistent with biological neural systems. This study calculates the accuracy rate without compromising the NNs' structure as a neural system mimic.

Increasing the number of hidden layers reduces the accuracy rate (this cause will be discussed in later articles 4.1). This result suggests that NNs-based AI does not necessarily produce better results with more hidden layers. As additional information, a correct response rate of approximately 80% is obtained when the recognition problem is solved by applying the correct response rate calculation method of this study without passing the MNIST data through the NNs.

The higher accuracy rate when expanding the range of correct answers indicates that even if the NNs cannot provide a completely correct answer, there is a high probability that the correct answer is among the "confused" options. This suggests that NNs' recognition and classification are not based on a binary logic of being correct or incorrect. In other words, NNs can recognize multiple possibilities, which ensures the versatile utility of deep learning artificial intelligence.

## 4. Discussions

### 4.1 Factors Contributing to the Decrease in Accuracy with Increased Hidden Layers

The reason for the decrease in accuracy as the number of hidden layers increases is considered by calculating the variance of the output vectors. It has been shown that calculating the variance of NNs' output vectors is an indicator of how much feature information is contained in the vectors [e.g., 24]. The larger the variance, the more feature information is included, and the smaller the variance, the less feature information is included, and the data is smoothed. In this study, the average variance of the calculated output vectors was computed for each number of hidden layers (Figure 2).

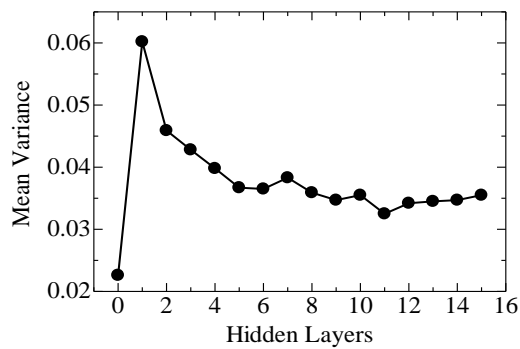

Figure 2: Average variance of output vectors for different numbers of hidden layers.

The variance is smallest when there are no hidden layers, i.e., when the input layer is fully connected to the output layer. This is likely because, in the absence of hidden layers, there is no activation function structurally present to contribute to the variance of the output vectors. When hidden layers are present, the variance decreases as the number of hidden layers increases up to around 10 layers. This indicates that the output vectors become smoothed as the number of hidden layers increases. The accuracy also decreases as the number of hidden layers increases, suggesting that the amount of variance decrease might contain the features necessary for the NNs and evaluation function in this study to make correct predictions. For more than 10 layers, the variance is about 0.035, which does not change even if the number of layers is increased to 15 or more. The average variance for 1 and 2 hidden layers differs by about 0.015, but there is almost no difference in accuracy. This suggests that there may be features in the data that cannot be analyzed by simple variance calculations, or that the evaluation function used in this study may not be able to calculate an accuracy rate higher than 80%.

**4.2 Improvements and Practical Applications of Feedforward NNs**

This study deal with a simple NNs structure to observe the primary information propagation capability of NNs. Like traditional NNs-based artificial intelligence, there is room for necessary improvements in layer structure, activation functions, and other information processing structures. These improvements could include convolutional NNs [25, 26], spiking NNs [27], and attention mechanisms [28]. The method for calculating accuracy can also be improved. This study used a comparison of the average vectors of training data and output vectors of test data, inspired by distributed representation. Other methods, such as creating multiple average vectors for each class, can also be considered. The average vector can be recalculated each time training data is input, and if a significant deviation from the average vector is observed in the output vector, the new output vector can be used as a new average vector, and both vectors can be used for validation.

This study only demonstrates the basic information propagation capability of NNs. In order to create practical software using feed-forward NNs without statistical weight optimization, it is essential to structure and algorithmize the NNs. There should be various paths to practical applications, ultimately aiming for a structure that mimics biological neural systems. The biological information processing algorithm operates in a cycle of Input → Processing → Output → Environmental Response → Input → ..., and this mimicry should be considered. This study focuses on input-related research. Creating an algorithm with a cyclical structure where processed information is output to the environment based on input information, and the impact of this output on the environment is re-input, can be an objective for functionalization. Processing includes recalling memories and adjusting to various input information based on emotions, requiring an approach that applies neuroscience. Output involves handling information that leads to actions, and connection with algorithms or robotics capable of executing these actions will be necessary. Additionally, developing an input mechanism capable of fully handling information from the environment is essential.

**5. Conclusion**

This study demonstrates that NNs could solve the MNIST classification problem with high accuracy without using statistical optimization such as error backpropagation by employing a simple feedforward NNs structure and distributed representation-based vector comparison. This finding highlights the primary information propagation capability of NNs. It is hoped that this approach will serve as a stepping stone toward constructing NNs-based artificial intelligence without mechanisms inconsistent with biological neural systems, ultimately contributing to the development of general-purpose artificial intelligence. The study observed that input data becomes smoothed as it passes through the hidden layers of NNs, and the loss of feature information is discussed through the calculation of the variance of output vectors.